%% file: main.tex
\documentclass[runningheads]{llncs}
\usepackage{booktabs} 
\usepackage[ruled,linesnumbered]{algorithm2e}

\usepackage{graphicx}
\usepackage{subcaption}
\usepackage{multirow}
\usepackage{csquotes}
\usepackage{paralist}
\usepackage{enumitem}

\newcommand\modelOne{\textsc{Reset\&FullTrain}}
\newcommand\modelTwo{\textsc{Reset\&CorrectedTrain}}
\newcommand\modelThree{\textsc{Accum\&FullTrain}}
\newcommand\modelFour{\textsc{Accum\&CorrectedTrain}}

\usepackage{xcolor}

\newcommand\accum{accumulative}
\newcommand\reset{reset}

\begin{document}
\title{Fairness-enhancing interventions in stream classification}
\author{Vasileios Iosifidis\inst{1,2}\and
Thi Ngoc Han Tran\inst{1}\and
Ntoutsi Eirini\inst{1,2}}
\authorrunning{Iosifidis et al.}
%
\institute{Leibniz University, Hannover, Germany \and
L3S Research Center, Hannover, Germany\\
\email{\{iosifidis,ntoutsi\}@L3S.de, tranthingochan.03@gmail.com}}

\maketitle

\begin{abstract}
The wide spread usage of automated data-driven decision support systems has raised a lot of concerns regarding accountability and fairness of the employed models in the absence of human supervision. 
Existing fairness-aware approaches tackle fairness as a batch learning problem and aim at learning a fair model which can then be applied to future instances of the problem. In many applications, however, the data comes sequentially and its characteristics might evolve with time. In such a setting, it is counter-intuitive to ``fix'' a (fair) model over the data stream as changes in the data might incur changes in the underlying model therefore, affecting its fairness. 
In this work, we propose fairness-enhancing interventions that modify the input data so that the outcome of any stream classifier applied to that data will be fair. Experiments on real and synthetic data show that our approach achieves good predictive performance and low discrimination scores over the course of the stream.

\end{abstract}
\section{Introduction}
\label{sec:indroduction}
\input{introduction}

\section{Related Work}
\label{sec:relatedwork}
\input{related}

\section{Fairness-aware stream classification}
\label{sec:ouralgorithms}
\input{methodologies}

\section{Experiments}
\label{sec:experiments}
\input{experiments}
 
\section{Conclusions and Future Work}
\label{sec:conclusions}
\input{conclusion}

\section*{Acknowledgements}
The work is inspired by the German Research Foundation (DFG) project OSCAR (Opinion Stream Classification with Ensembles and Active leaRners) for which the last author is Co-Principal Investigator.

\bibliographystyle{splncs04}
\bibliography{bibliography}
\end{document}

%% file: introduction.tex

Despite the wide spread belief that data-driven decision making is objective in contrast to human-based decision making that is subject to biases and prejudices, several cases have been documented, e.g.,~\cite{airbnb,AmazonPrime}, in which data-driven decision making incurs discrimination. As a recent example, a Bloomberg report has suggested signs of racial discrimination in Amazon's same-day delivery service~\cite{AmazonPrime}. The sensitive attribute race was not employed as a predictive attribute in Amazon's model(s), however the location of the users might have acted as a proxy for race. As a result, predominantly black ZIP codes were excluded from services and amenities. 
Therefore, the wide spread usage of automated data-driven decision support systems has raised a lot of concerns regarding accountability and fairness of the employed models in the absence of human supervision~\cite{bhandari2016big,united2014big,sweeney2013discrimination}.
Such issues result in societal and legal implications, therefore, recently the domain of discrimination-aware data mining~\cite{pedreshi2008discrimination} has attracted a lot of attention and several methods have been proposed ranging from discrimination discovery to discrimination elimination and explanation of model decisions.

Most of these methods, however, tackle fairness as a batch learning problem aiming at learning a ``fair'' model which can be then used for predicting future instances of the population.
In many modern applications, however, data is generated sequentially and its characteristics might change with time, i.e., the data is non-stationary. Such dynamic environments( or, data streams) call for model adaptation~\cite{gama2010knowledge}.
As an example, in the EU, the non-native population has significantly changed in the last years due to European refugee crisis and internal EU migration with a potential effect on the racial discrimination in the labor market.
In such \emph{non-stationary} environments, the main challenge for supervised learning is the so-called concept drifts, i.e., changes in the underlying data distribution which affect the learning model as the relationships between input and class variables might evolve with time~\cite{schlimmer1986beyond}. Existing solutions from the data stream mining domain tackle this issue by adapting the learning models online. 
However, as the decision boundary of the classifier changes as a result of model adaptation, the fairness of the model might get hurt.

\begin{figure}[t!]
\centering
 \includegraphics[width=1\textwidth]{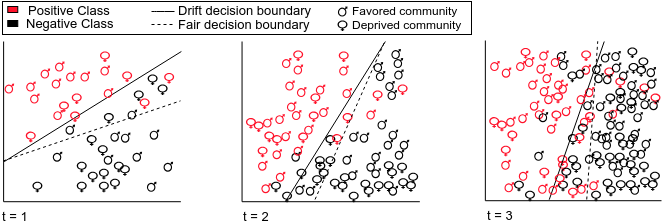}
 \caption{An evolving stream with discrimination: At each timepoint, the current decision boundary (solid line) and the ``fairness-corrected'' boundary (dashed line) are displayed.}
 \label{fig:discriminationStream}
\end{figure}

An example of an evolving stream with discrimination is shown in Figure~\ref{fig:discriminationStream}; one can see the deprived and favored communities (w.r.t. some sensitive attribute) over time as well as their class assignments. The favored community dominates the stream.
The decision boundary of the classifier (solid line) changes in response to changes in the underlying data. As a result, the associated fairness of the model also changes, calling for ``fairness-enhancing intereventions'' (dashed line).
It is important, therefore, model adaptation to also consider fairness` to ensure that a valid fairness-aware classifier is maintained over the stream. 
In this work, we propose fairness-enhancing interventions that modify the input data before updating the classifier. Our method belongs to the category of pre-processing approaches to fairness, investigated that far only in the context of static learning~\cite{calders2009building,calmon2017optimized,iosifidisdealing,kamiran2009classifying,kamiran2010classification,kamiran2012data}.
Our contributions are:
\begin{inparaenum}[a)]
	\item we introduce the fairness-aware classification problem for streams
	\item we propose pre-processing
	fairness-enhancing interventions for streams
	\item we propose a synthetic generator for simulating different drift and fairness behaviors in a stream 
	and
	\item we present an extensive experimental evaluation with different stream learners and on different datasets.
\end{inparaenum}

The rest of the paper is as follows: In Section 2, we overview the related work. Our approach is presented in Section~\ref{sec:ouralgorithms}. Experimental results are discussed in Section~\ref{sec:experiments}. Finally, conclusions and outlook are presented in Section~\ref{sec:conclusions}. 
 
%% file: related.tex
Although more than twenty different notions of fairness have been proposed in the last few years~\cite{romei2014multidisciplinary,verma2018fairness}, still there is no agreement on which measure to apply in each situation.
The most popular is that of statistical parity~\cite{verma2018fairness} that checks whether the favored and deprived communities
have equal probability of being assigned to the positive class. This is the measure we also adopt in this work.

\textbf{Pre-processing fairness-enhancing interventions:} 
Methods in this category work under the assumption that in order to learn a fair classifier, the training data should be discrimination-free. To this end, they try to balance the representation of the different groups in the population. 
Massaging~\cite{kamiran2009classifying} modifies the data distribution by re-labeling some of the instances which reside close to the decision boundary in order to neutralize discriminatory effects.
Re-weighting~\cite{calders2009building} assigns different weights to the different group, e.g., the deprived group will receive a higher score comparing to the favored group. 
These methods are typically model-agnostic and therefore, any classifier is applicable after the pre-processing phase.

\textbf{In-processing fairness-enhancing interventions:} Methods in this category directly modify the learning algorithm to ensure that it will produce fair results. As such, they are algorithm-specific; e.g.,~\cite{kamiran2010discrimination} proposes a decision tree that encodes fairness by employing a modified entropy-based attribute splitting criterion and~\cite{dwork2018decoupled} 
includes sensitive attributes in the learning
process by utilizing a joint loss function that makes explicit trade-off between fairness and accuracy.

\textbf{Post-processing fairness-enhancing interventions:} Post-processing methods modify the results of a trained classifier to ensure the chosen fairness criterion is met; e.g.,~\cite{kamiran2010discrimination} modifies the leaf labels of a decision tree, ~\cite{pedreschi2009measuring} changes the confidence values of classification rules and~\cite{fish2016confidence} shifts the decision boundary of an AdaBoost learner until the fairness criterion is fulfilled. 

\textbf{Stream Classification:} 
Data stream algorithms must be able to adapt to concept drifts in order to maintain a good performance over the stream~\cite{gama2010knowledge}. Model adaptation is typically enabled by: i) incorporating new instances from the stream into the model and ii) forgetting or downgrading outdated information from the model. The former calls for online/incremental algorithms, whereas the latter calls for methods that are able to forget e.g., \cite{forman2006tackling,klinkenberg2004learning}. We discuss several stream classifiers in the experiments (Section~\ref{sec:experiments}).

\textbf{Sequential fairness:} When a sequence of decisions has to be taken, the notion of sequential fairness is relevant. For example,~\cite{stoyanovich2018online}, studies fair online item ranking for groups and~\cite{liu2018delayed} how fairness criteria interact with temporal indicators of well-being and affect discriminated populations on the long-term.

Our work lies in the intersection of pre-processing methods for fairness and stream classification methods. The former, however, focus solely on the static case, i.e., they assume that the data is stationary, whereas the latter focus solely on predictive accuracy and ignore fairness.
To the best of our knowledge, this is the first work trying to bridge the two domains.

%% file: methodologies.tex
A data stream $S$ 
is a potentially infinite sequence of instances arriving over time, each instance described in a feature space $A=(A_1, A_2 \cdots A_d)$. 
One of the attributes is the \emph{sensitive attribute}, denoted by $SA$, with values $SA=\{s,\overline{s}\}$; we refer to $s$ and $\overline{s}$ as ``deprived'' and ``favored'', respectively. 
 We also assume a binary class attribute $C=\{rejected, granted\}$. We refer to \enquote{granted} class value as \emph{target class}. 
We process the stream in chunks of fixed size, $S_1, \cdots, S_t$ with $S_t$ being the most recent chunk. We assume the fully supervised learning setting, where the labels of the instances are available shortly after their arrival.
Therefore, the goal is to make a prediction for the instances based on the current classifier and use the labels later on for update (the so-called prequential evaluation~\cite{gama2010knowledge}).
The underlying stream population is subject to changes, which might incur concept drifts, i.e., the decision boundary might change overtime (c.f., solid line in Figure~\ref{fig:discriminationStream}) and therefore, fairness implications may take place (c.f., dashed line in Figure~\ref{fig:discriminationStream}). A stream classifier typically takes care of concept drifts, but does not consider fairness. 

\begin{figure}[ht!]
	\centering
	\includegraphics[width=1\linewidth]{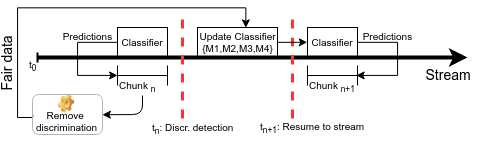}
	\caption{Fairness-aware stream classification overview}
	\label{fig:architecture}
\end{figure}

The \emph{discrimination aware stream classification problem} therefore is to maintain a classifier that performs well (i.e., the predictive accuracy is high) and does not discriminate (i.e., the discrimination score is low, c.f. Equation~\ref{eq:stream_disc}) over the course of the stream. 
In this work, we follow the pre-processing approaches to fairness-aware learning that intervene at the input data to ensure a fair representation of the different communities.
In particular, we monitor the discrimination in each incoming chunk from the stream (Section~\ref{sec:discriminationScoreStream}) and if the discrimination score exceeds a user defined threshold $\epsilon$, we \enquote{correct} the chunk for fairness (Section~\ref{sec:discriminationCorrectionStream}) before feeding it into the learner (Section~\ref{sec:discriminationModelUpdateStream}).
We assume an initialization phase at the beginning of the stream for which an initial fairness-aware classifier $F_0$ is trained upon an initial dataset $S_0$ from the stream.
An overview of our approach is depicted in Figure~\ref{fig:architecture}, where M1-M4 are the adaptation strategies introduced in Section~\ref{sec:discriminationModelUpdateStream}.

\subsection{Detecting classifier discrimination in data streams}
\label{sec:discriminationScoreStream}
\input{stream_discrimination}

\subsection{Fairness-enhancing data interventions in data streams}
\label{sec:discriminationCorrectionStream}
If discrimination is detected, $S_t$ is ``corrected'' for fairness before being used for model update (Section~\ref{sec:discriminationModelUpdateStream}). To this end, we employ two different data intervention techniques: massaging and re-weighting.

\textbf{Chunk-based massaging.}
\input{stream_massaging}
\textbf{Chunk-based re-weighting.}
\input{stream_reweighting}

\textbf{Massaging vs re-weighting.}
Both methods modify the data distribution to equalize the number of deprived and favored communities in the target class.
However, there are fundamental differences between the two approaches: massaging interferes at the instance level by altering single instances, whereas re-weighting affects a whole community by lowering/increasing its weight. Moreover, massaging is more intrusive than re-weighting as it alters the class labels. 
Both interventions result in a \enquote{corrected} chunk $S_t'$ $(|S_t|=|S_t'|)$ used for updating the classifier (c.f., Section~\ref{sec:discriminationModelUpdateStream}).

\subsection{Fairness-aware classifier adaptation in data streams}
\label{sec:discriminationModelUpdateStream}
\input{strategies}

%% file: stream_discrimination.tex
Let $F$ be the current (non-discriminating) stream classifier and $S_t$ be the current chunk received from the stream. 
We evaluate the discriminative behavior of $F$ over $S_t$, i.e., $disc_S(F,S_t)$ by evaluating the predictions of $F$ over instances of $S_t$. 
First, we define four communities in each chunk $S_t$ by combining the sensitive attribute $SA$ with the (predicted) class attribute $C$ (both binary attributes):
\begin{table}[ht]\small
	\centering
		\caption{(Chunk $S_t$) Communities}
	\label{tab:confusionmatrix}
	\begin{tabular}{ccc}
		\hline
		& \multicolumn{2}{c}{\textbf{(predicted) class}} \\
		\begin{tabular}[c]{@{}c@{}}\textbf{Sensitive Attribute SA}\end{tabular} & Rejected & Granted \\ \hline
		$s$ (Female) & $DR_t$ \textit{(deprived rejected)} & $DG_t$ \textit{(deprived granted)} \\
		$\overline{s}$ (Male) & $FR_t$ \textit{(favored rejected)} & $FG_t$ \textit{(favored granted)} \\ \hline
	\end{tabular}
\end{table}\normalsize

As discrimination measure, we employ statistical parity that evaluates whether the favored and deprived groups have equal probabilities of being granted~\cite{romei2014multidisciplinary}:

\begin{equation}
\label{eq:stream_disc}
disc_S(F,S_t) = \frac{FG_t}{FG_t + FR_t} - \frac{DG_t}{DG_t + DR_t} 
\end{equation}
If the discrimination value exceeds the threshold $\epsilon$, i.e., $disc_S(F,S_t) > \epsilon$, the discrimination performance of the model degrades, due to, e.g., changes in the distribution that reside in the newly arrived chunk $S_t$.
A typical stream classifier would update $F$ based on $S_t$ to adapt to the incoming stream. However, to also account for fairness, we first ``correct'' $S_t$ for fairness (c.f., Section~\ref{sec:discriminationCorrectionStream}), before employing its instances for updating the model (c.f., Section~\ref{sec:discriminationModelUpdateStream}).



%% file: stream_massaging.tex
Massaging~\cite{kamiran2009classifying} modifies the data distribution by swapping the class labels of certain instances (from \enquote{granted} into \enquote{rejected} or vise versa) from each of the deprived rejected (DR) and favoured granted (FG) communities. 
The amount of affected instances, $M_t$, from each community is derived by Equation~\ref{eq:stream_disc} and is as follows:

\begin{equation}	
\label{eq:M_score_for_stream}
M_t = \frac{FG_{S_t} * (DG_{S_t} + DR_{S_t}) - DG_{S_t} * (FG_{S_t} + FR_{S_t})}{|S_t|}
\end{equation}
The best candidate instances for label swapping are those close to the decision boundary, as intuitively their alternation will have the least impact on the model while it will fulfill the discrimination requirement (Equation~\ref{eq:stream_disc}). To this end, we employ a ranker $R_t$ trained on $S_t$ that estimates the class probabilities of the instances in $S_t$. Then, $M_t$ instances assigned with high probability to $DR$ and $M_t$ instances assigned with low probability to $FG$ are selected for label swapping.

%% file: stream_reweighting.tex
Re-weighting~\cite{calders2009building} modifies the data distribution by assigning different weights to each community (c.f., Table~\ref{tab:confusionmatrix}) to \enquote{enforce} a fair allocation of deprived and favored instances w.r.t the target class. 
Similarly to massaging, the exact weights depend on the $S_t$ distribution in the different communities.
Below we provide the weight for the favoured granted community, same rationale holds for the other communities:
\begin{equation}\label{eq:static_weighting}
W_{t}^{FG} = \frac{| \overline{s}_{S_t}| * |\{ x \in S_t |(x.C = ``granted")\}|}{|S_t| * |FG_{S_t}|} 
\end{equation}
Each instance $x \in S_t$ is weighted by \enquote{inheriting} the weight of its community. 



%% file: strategies.tex
The update of a classifier should take into account both concept drifts and fairness. For the former, we work with stream classifiers, like, Hoefdding Trees, Accuracy Updated Ensembles and Naive Bayes that already adapt to concept drifts. In that sense, the concept drift problem is directly tackled by the learner. 
For the latter, we ``correct'' the input stream per chunk, using either massaging or re-weighting, to ensure that learners are trained on ``fair'' data (c.Section~\ref{sec:discriminationCorrectionStream}).
In particular, we propose update strategies for fairness-aware stream classification:
\begin{itemize}[leftmargin=*]
	\item \modelThree~(shortly $M1$): $F$ is continuously updated over the stream using the original current chunk $S_t$, if no discrimination is detected, or its \enquote{corrected} counterpart $S_t'$, if discrimination is detected.		
	\item\modelOne~(shortly $M2$): Similar to M1, but if discrimination is detected, $F$ is reset and a new model is created from the ``corrected'' $S_t'$.
\end{itemize}

The underlying assumption for~\modelThree~is that if trained with ``fair'' chunks, the classifier $F$ should be fair. In practice, though and due to the complex interaction between input data and learning algorithms, this might not be true (c.f., \cite{kamiran2012data}); therefore, we also propose the~\modelOne~model that resets the learner once its predictions incur discrimination.

In addition, we propose two variations that focus more on fairness. The rationale is similar to the previous approaches but the model is updated only if discrimination is detected and only via \enquote{corrected} data. Therefore, these two models adapt slower to concept drifts comparing to the first two models, as their adaptation occurs only if discrimination is detected.
In particular:
\begin{itemize}
\item\modelFour~(shortly $M3$): $F$ is updated over the stream only if discrimination is detected. Update is based on \enquote{corrected} chunks $S_t'$. 
	\item\modelTwo~(shortly $M4$): Similar to M3, but once discrimination is detected $F$ is reset and a new model is created from the corrected chunk $S_t'$. Thus, $M4$ adapts only via reset, when discrimination is detected.
\end{itemize}


%% file: experiments.tex
We evaluate the performance of our methods for discrimination elimination in data streams using both real and synthetic datasets (Section~\ref{sec:datasets}).
As evaluation measures, we use the performance of the model, in terms of accuracy and discrimination, over the new coming chunk from the stream.
We report on the performance of the different methods over the stream but also on the overall accuracy-vs-fairness behavior of the different methods.
We experiment with a variety of stream classifiers such as Naive Bayes (NB), Hoeffding Tree (HT), Accuracy Updated Ensemble (AUE) and k-Nearest Neighbors (KNN). 
The aforementioned models are updated based on the new incoming chunk from the stream, however they differ w.r.t how they handle historical information. NB and HT classifiers do not forget, whereas AUE forgets by replacing old learners with new ones. kNNs on the other hand, rely solely on the last chunk for the predictions, due to its internal buffer. An overview of each classifier is given below:
\begin{itemize}[leftmargin=*]
\item Naive Bayes (NB): A probabilistic classifier that makes a simplistic assumption on the class-conditional independence of the attributes. 
The stream version of NBs~\cite{bifet2010sentiment} is an online algorithm, i.e., the model is updated based on new instances from the stream, but does not forget historical information.

\item Hoeffding Tree (HT): A decision tree classifier for streams that uses the Hoeffding bound to make a reliable decision on the best splitting attribute from a small data sample~\cite{domingos2000mining}. HT is an online algorithm (so, it is updated based on new instances from the stream) but does not forget.

\item Accuracy Updated Ensemble (AUE): An ensemble model that adapts to concept drifts by updating its base-learners based on the current
data distribution, tuning their weights according to their predictive power on the most recent chunk~\cite{brzezinski2011accuracy}. The model replaces old learners with newer ones trained upon more recent chunks. We used HTs as base learners for the ensemble and we set the maximum number of base learners to 10.

\item KNN: A lazy learner which predicts based on the class labels of the neighboring instances~\cite{bifet2010moa}. In particular, the previous chunk instances and their labels are used to make predictions for the instances of the current chunk. The neighborhood is set to $k=10$. 
\end{itemize}


We evaluate our strategies M1-M4 (c.f., Section~\ref{sec:discriminationModelUpdateStream}) for the different classifiers, as well as againsts the following baselines that do not explicitly handle discrimination:
\begin{itemize}
	\item[B1] \textbf{\textsc{B.NoSA}} (Baseline NoSensitiveAttribute): The classifier $F$ does not employ $SA$ neither in training nor in testing. The model is continuously updated over the stream from the original chunks $S_t$. Intuitively, the model tackles discrimination by omitting $SA$.
	\item[B2] \textbf{\textsc{B.RESET}} (Baseline Reset): If discrimination is detected, the old model $F$ is deleted and a new model is learned on $S_t$. The model is updated over the stream, but without any correction. Discrimination is being monitored and if it is detected again, the whole procedure starts over. Intuitively, this approach tackles discrimination by resetting the model when discrimination is detected.
\end{itemize}

For the massaging techniques, we use NB as a ranker which according to~\cite{calders2009building} is the best ranker. We implemented our methods\footnote{Code will be made available online} in MOA~\cite{bifet2010moa}. 
For all of our reported experiments, we consider a discrimination threshold of $\epsilon = 0.0$, that is, we do not tolerate any discrimination, and a chunk size of $|S| = 1,000$ instances. The effect of these parameters is discussed in Section~\ref{sec:experiments_eval}.

\subsection{Datasets}
\label{sec:datasets}
\input{datasets}

\subsection{Evaluation results}
\label{sec:experiments_eval}
\begin{figure*}[t!]
 \centering
 \begin{subfigure}[b]{0.7\textwidth}
 \centering
 \includegraphics[width=\textwidth]{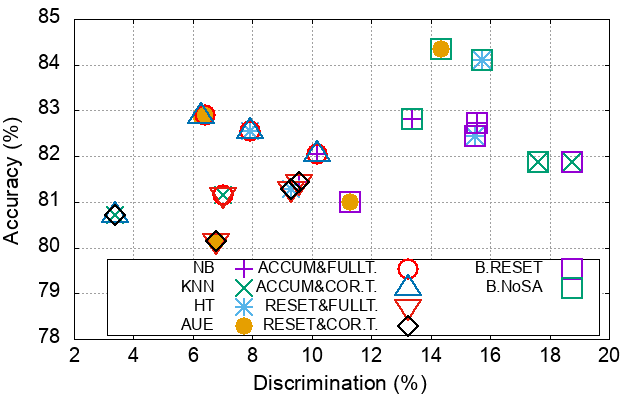}
 \caption[]%
 {{\small Massaging}} 
 	\label{fig:small_msg_withSA}
 \end{subfigure}
 \hspace{-12.5pt}
 \begin{subfigure}[b]{0.7\textwidth} 
 \centering 
 \includegraphics[width=\textwidth]{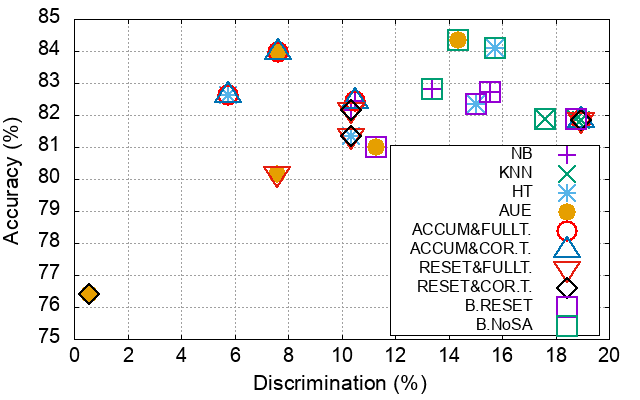}
 \caption[]%
 {{\small Re-weighting}} 
 	\label{fig:small_overall_disc_acc_withSA_rw}
 \end{subfigure}
 \caption{Census-Income (Small): Discrimination-vs-Accuracy of the different strategies}
 \label{fig:overall_census}
\end{figure*}

For each dataset, we report on the discrimination-vs-accuracy behavior of the different classifiers under the different adaptation strategies. The discrimination-vs-accuracy plot (an example is shown in Figure~\ref{fig:small_msg_withSA}) allows for a quick evaluation of the different behaviors. 
Values close to 0 in the $x$-axis mean fair models, whereas as the values increase the corresponding classifier become more discriminating. w.r.t accuracy ($y$-axis), good predictive power models reside close to 100\%, whereas low $y$ values indicate poor performing models.
The ideal models are located on the up left region which indicates high accuracy and low discrimination performance models. The worst models are located in the bottom right region where low accuracy and high discriminating behavior take place. Up right and bottom left regions indicate unfair but accurate models and fair but inaccurate models, respectively.
\begin{figure}[t!]
	\centering	
	\includegraphics[width=.82\linewidth]{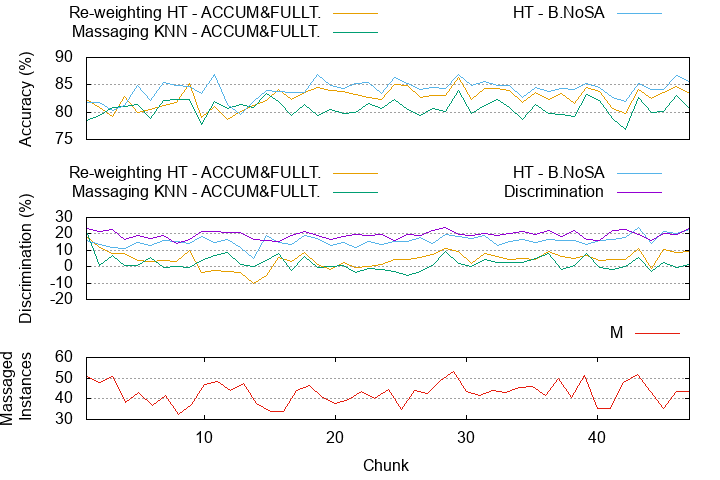}
	\caption{Census-Income: Accuracy (top), discrimination (middle) and \# massaged instances (bottom) over the stream}
	\label{fig:small_stream_comparison}
\end{figure}

\textbf{Census-Income.} For the massaging, c.f., Figure~\ref{fig:small_msg_withSA}, our strategies achieve lower discrimination comparing to the baselines (our values are closer to 0 in the $x$-axis). As expected, the improvement w.r.t discrimination incurs a drop in accuracy, i.e., baselines have better accuracy comparing to our strategies (baseline values are closer to 100\% in the $y$-axis). 
We also observe that some strategies depict very similar performance, e.g., $M2$ and $M4$ when combined with $HT$. The reason is that since $\epsilon=0$, our discrimination detector is activated on almost every chunk from the stream and therefore strategies like $M2$ and $M4$ will both reset the model on each chunk.
Accumulative strategies, $M1$  and $M3$, perform better than \reset~strategies, $M2$ and $M4$; the reason is probably that the latter ones forget too fast. 
Regarding the different classifiers employed by our strategies, we can see that the best performing ones in terms of both accuracy and discrimination are $KNN$ and $AUE$. 
$AUE$ and $HT$ models yield better accuracy and less discrimination when they do not discard previous knowledge. 
Although $KNN$ is not the best performing model in terms of accuracy, it yields the lowest discrimination score with the smallest drop in accuracy when compared to its baselines, namely $B.RESET$ and $B.NoSA$. In particular, discrimination drops from 19\% to 4\% while accuracy drops by almost 1\%, when $KNN$ is employed by $M3$ and $M4$ strategies. 

In Figure~\ref{fig:small_overall_disc_acc_withSA_rw}, we compare the discrimination-vs-accuracy behavior of the different classifiers under re-weighting. Same as in massaging, our strategies reduce discrimination in predictions. Classifiers such as HT behave similarly under different strategies since the detector detects discrimination in almost every chunk. KNN on the other hand, doesn't take into consideration weights, hence all the strategies perform identically. 

We also compare models overtime in Figure \ref{fig:small_stream_comparison}. We have selected one model for each method (massaging/re-weighting) based on the discrimination-accuracy trade off (points which are closer to (0,1) based on Euclidean distance) and the best baseline of those two models. Although HT's baseline has the best accuracy overtime, its discrimination score is close to stream's discrimination. On the other hand, re-weighting and massaging methods result in a significant drop in discrimination. The number of massaged instances $M$ varies over the stream, based on how discriminative the KNN's predictions are.
\begin{figure*}[t!]
 \centering
 \begin{subfigure}[b]{0.7\textwidth}
 \centering
 \includegraphics[width=\textwidth]{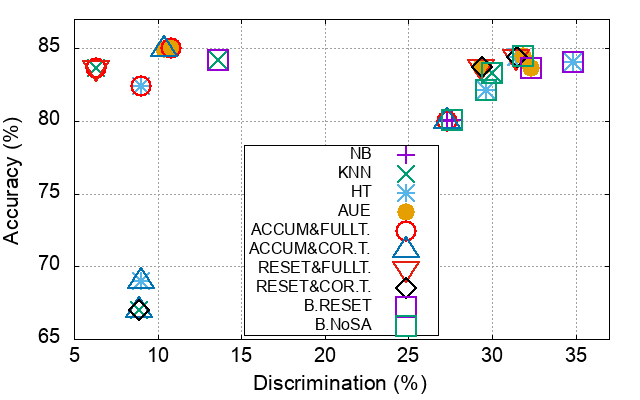}
 \caption[]%
 {{\small Massaging}} 
	\label{fig:stream_overall_disc_acc_withSA}
 \end{subfigure}
 \hspace{-12.5pt}
 \begin{subfigure}[b]{0.7\textwidth} 
 \centering 
 \includegraphics[width=\textwidth]{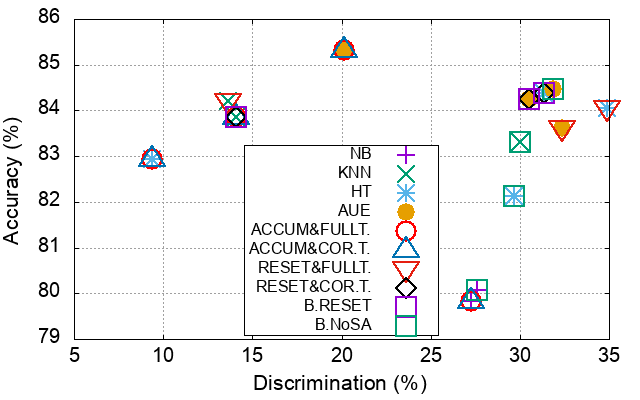}
 \caption[]%
 {{\small Re-weighting}} 
	\label{fig:stream_overall_disc_acc_withSA_rw}
 \end{subfigure}
 \caption{Synthetic stream: Discrimination-vs-Accuracy of the different strategies}
 \label{fig:synthetic_results}
\end{figure*}
\begin{figure}[t!]
	\centering
	\includegraphics[width=.82\linewidth]{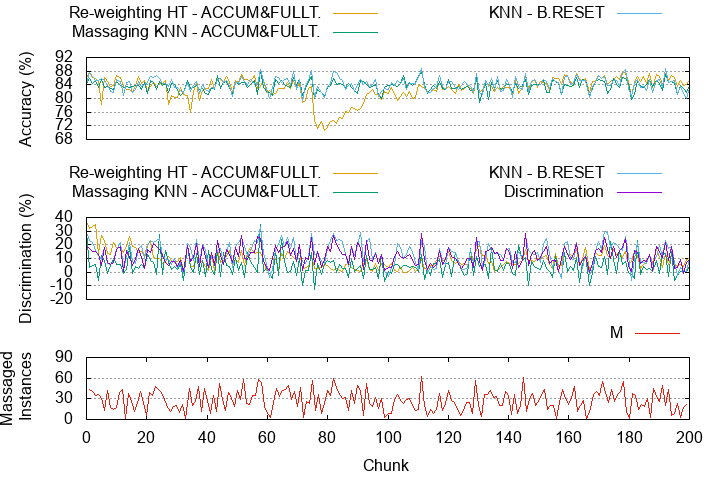}
	\caption{Synthetic stream: Accuracy (top), discrimination (middle) and \# massaged instances (bottom) over the stream}
	\label{fig:synthetic_stream_comparison}
\end{figure}

\textbf{Synthetic data.}
The dataset contains 20 concept drifts while its discrimination score varies overtime, as seen in Figure \ref{fig:streamBasic}. The majority of baselines, in Figures \ref{fig:stream_overall_disc_acc_withSA} and \ref{fig:stream_overall_disc_acc_withSA_rw}, are able to adapt to concept drifts (i.e., they achieve high accuracy), however they cannot handle discrimination, which in some cases is even amplified comparing to the original stream overall discrimination.
The vast majority of baselines occupy the up right region which means that models are able to adapt to concept drifts even though they are highly discriminating. 
By inspecting Figure~\ref{fig:stream_overall_disc_acc_withSA}, we can observe once again that \accum~models are less discriminating in contrast to \reset~models. KNN achieves high reduction in discrimination (up to 6\%), while maintaining high accuracy. Classifiers such as AUE and HT perform well when combined with \accum~strategies while \reset~strategies incur higher discrimination.
A possible reason for this behavior is that when trained on more data, a model can generalize better, especially in re-occurring concepts, comparing to reset strategies that rely solely on recent chunks.
KNN is an exception as it performs well despite relying on an internal sliding window for its predictions. A possible reason is that a kNN learner is an instance-based learned and does not perform explicit generalization like HT and AUE. Similarly to census-income dataset, NB is not able to tackle discrimination.

In Figure~\ref{fig:stream_overall_disc_acc_withSA_rw}, we observe that almost all baselines, same as in massaging, cover the up right region area. AUE's performance is increasing while it becomes more discriminating in contrast to AUE in massaging. Same as before, HT and KNN have the least discriminating behavior while NB performs poorly. Again, \reset~strategies produce good accuracy models but fail to reduce discrimination. 

In Figure \ref{fig:synthetic_stream_comparison}, we compare the best ``correction'' methods and the best baseline. $M1$ combined with KNN has the lowest discrimination score overtime. Its accuracy is slightly worse than its baseline. However, discrimination is lower than stream's and baseline's discrimination overtime. HT's overall performance w.r.t accuracy is relatively good except the interval between 77th and 90th chunk where four concept drifts occurred incurring accuracy loss. Despite the accuracy degradation, HT achieved lower discrimination compared to other classifiers.

\textbf{Parameter effect:}
Due to lack of space we omit the time execution charts. A derived conclusion is that our strategies are executed slightly slower compared to the baselines and moreover, that the reset strategies are faster than the accumulative strategies. 
We have also experimented with different chunk sizes $|S|$ and discrimination thresholds $\epsilon$. Based on our experiments, increasing $\epsilon$ results in better accuracy models but their discrimination also increases. 
With respect to the chunk size effect, there was no clear effect on the performance except for the execution time that decreases with chunk size as less operations take place.

%% file: datasets.tex
As real dataset we employ the census-income (or adult-census) dataset, which comprises one of the most popular datasets in this area; we simulate the stream using the file order. 
Due to lack of stream data for fairness, we extend an existing stream generator to simulate different discrimination scenarios in data streams.

\begin{figure*}[t!]
	\centering
	\medskip
	\begin{subfigure}[t]{.73\linewidth}
		\centering
		\includegraphics[width=\textwidth]{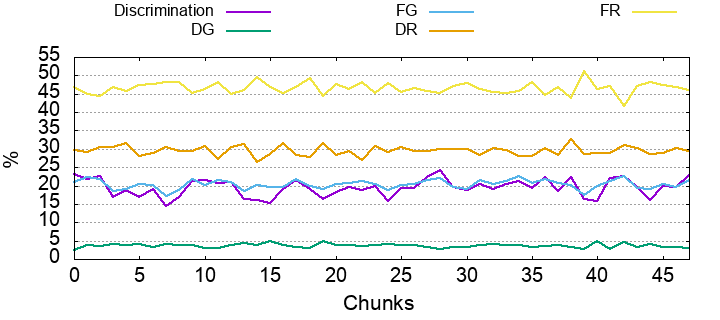}
		\caption{Census-Income}
		\label{fig:censusSmallBasic}
	\end{subfigure}
	\begin{subfigure}[t]{.73\linewidth}	
		\centering
		\includegraphics[width=\textwidth]{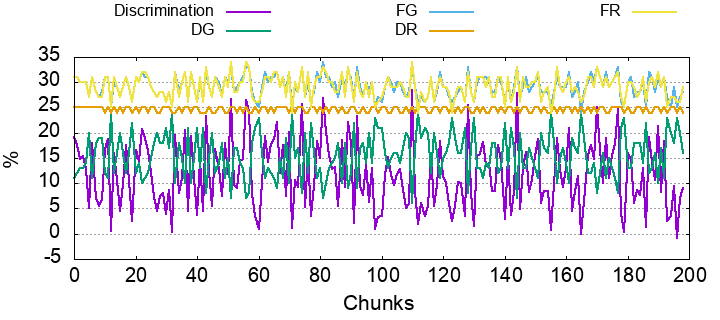}
		\caption{Synthetic data}
		\label{fig:streamBasic}
	\end{subfigure}
	~
	\caption{Discrimination and community size over the stream ($|S|=1,000$ instances/chunk)}
\end{figure*}

\textbf{Census-Income~\cite{merz1998uci}}:
\label{sec:datasets:censusSmall}
The learning task is to predict whether a person earns more than 50K/year using demographic features. We consider gender as the sensitive attribute with females being the deprived community and males being the favored community. In addition, we consider an annual income of more than 50K as the target class.
The dataset consists of 48,842 records and has an overall discrimination of 19.45\%. 
In Figure~\ref{fig:censusSmallBasic}, the discrimination score and the different community volumes ($DR_t, DG_t, FR_t, FG_t$) are shown over time using a chunk size of $|S|=1,000$ instances. The discrimination score ranges between $15\%-25\%$ overtime.

\textbf{Synthetic Generator:}
\label{sec:datasets:synthetic}
\input{synthetic_generator}

%% file: synthetic_generator.tex
Our generator comprises an extension of the static fairness generator of~\cite{zafar2017fairness} that represents each community using a Gaussian distribution. It forces the DG community to be closer to the negative class whereas the FG community is placed further away from the negative class. An example can be already seen in Figure~\ref{fig:discriminationStream}.
We extend this idea in a stream setting by varying the amount of discrimination over the stream while introducing concept drifts. 

In particular, we initialize four Gaussians, as follows, similarly to the static generator: \textit{p(DG) = N([2; 2], [3, 1; 1, 3])}, \textit{p(FG) = N([2.5; 2.5], [3, 1; 1, 3])}, \textit{p(DR) = N([0.5; 0.5], [3, 3; 1, 3])} and \textit{p(FR) = N([-2; -2], [3, 1; 1, 3])}.
In the initialization phase, all Gaussians contribute equally to each community with $n$ instances each, giving a total of $N=4n$ instances for the initial chunk.
With respect to discrimination, we introduce a parameter $SPP$ that controls the statistical parity by controlling the number of generated instances $x$ in the DG community over the stream.
The exact amount of instances $x$ can be derived from Equation \ref{eq:stream_disc} as follows:
\begin{equation}
 SPP = \frac{n}{2 * n} - \frac{x}{x + n} \Rightarrow x = n*\frac{1 - 2* SPP}{1 + 2 * SPP} 
\end{equation} 
where $n$ is the amount of instances generated by each Gaussian in a chunk and $x$ is the amount of instances for the $DG$ community based on the desired $SPP$ value; the rest $n-x$ instances generated originally by its corresponding Gaussian are evenly distributed to the FG and FR communities. This way, the ratio of positive instances in the favored community remains the same.
To simulate concept drifts in the population, we change the means of the Gaussians at random points over the stream. To maintain the initial problem (unfair treatment of one community), we shift the means all together at a random direction \textit{up, down, left or right} by a random value $k \in [0,2]$.

For evaluation purposes, we generate a synthetic dataset of 200,000 instances (200 chunks, $N=1,000$ instances per chunk), 4 numerical attributes, 1 binary sensitive attribute and 1 binary class.
We inserted 20 concept drift at random points and vary $SPP$ randomly over time from 0\% to 30\%. 
The dataset characteristics are shown in Figure \ref{fig:streamBasic}. 


%% file: conclusion.tex
In this work, we proposed an approach for fairness-aware stream classification, which is able to maintain good predictive performance models with low discrimination scores overtime. Our approach tackles discrimination by \enquote{correcting} the input stream w.r.t fairness and therefore, can be coupled with any stream classifier. Our experiments show that such a correction over the stream can reduce discrimination in model predictions, while the maintenance of the model over the stream allows for adaptation to underlying concept drifts.
Comparing the different fairness-intervention methods, our experiments show that massaging performs better than re-weighting. 
A possible explanation is that massaging works at an individual instance level by swapping its class label, whereas re-weighting works at a group level by applying different weights to different communities. Moreover, massaging affects selected instances, which are closer to the boundary. 

Our approach is model-agnostic, however our experiments show that the effect of \enquote{data correction for discrimination} on a variety of classifiers is different and therefore, how to \enquote{best correct} for specific classifiers is an interesting research direction. 
Moreover, we want to investigate in-processing fairness-aware stream classifiers that incorporate fairness notion directly in the classification algorithm.